\documentclass[10pt,twocolumn,letterpaper]{article}

\usepackage[pagenumbers]{cvpr} 

\definecolor{cvprblue}{rgb}{0.21,0.49,0.74}
\usepackage[pagebackref,breaklinks,colorlinks,allcolors=cvprblue]{hyperref}
\usepackage{xcolor}
\usepackage{booktabs}      
\usepackage[table]{xcolor} 
\usepackage{multirow}      
\usepackage{graphicx}      
\usepackage{pifont}
\usepackage{multirow}

\def\paperID{16660} 
\def\confName{CVPR}
\def\confYear{2026}

\newcommand{\fref}[1]{Fig.~\ref{#1}}

\newcommand{\tref}[1]{Tab.~\ref{#1}}

\title{$\mathbf{M^3A}$ Policy: Mutable Material Manipulation Augmentation Policy through Photometric Re-rendering}

\author{Jiayi Li$^{1,2,*}$, Yuxuan Hu$^{2}$, Haoran Geng$^{3}$, Xiangyu Chen$^{2}$, Chuhao Zhou$^{2}$,
\\Ziteng Cui$^{4}$ and Jianfei Yang$^{2,\dagger}$ \\
$^{1}$Tsinghua University \quad
$^{2}$MARS Lab, Nanyang Technological University
\\
$^{3}$University of California, Berkeley \quad
$^{4}$The University of Tokyo
\\
$^*$ Work carried out during NTU Research Internship\\
$^\dagger$Corresponding author
\\
{\tt\small jy-l21@mails.tsinghua.edu.cn, jianfei.yang@ntu.edu.sg}
}

\begin{document}
\twocolumn[
{
\renewcommand\twocolumn[1][]{#1}
\maketitle

\begin{center}
    \centering
    \vspace{-4mm}
    \captionsetup{type=figure}
    \includegraphics[width=\linewidth]{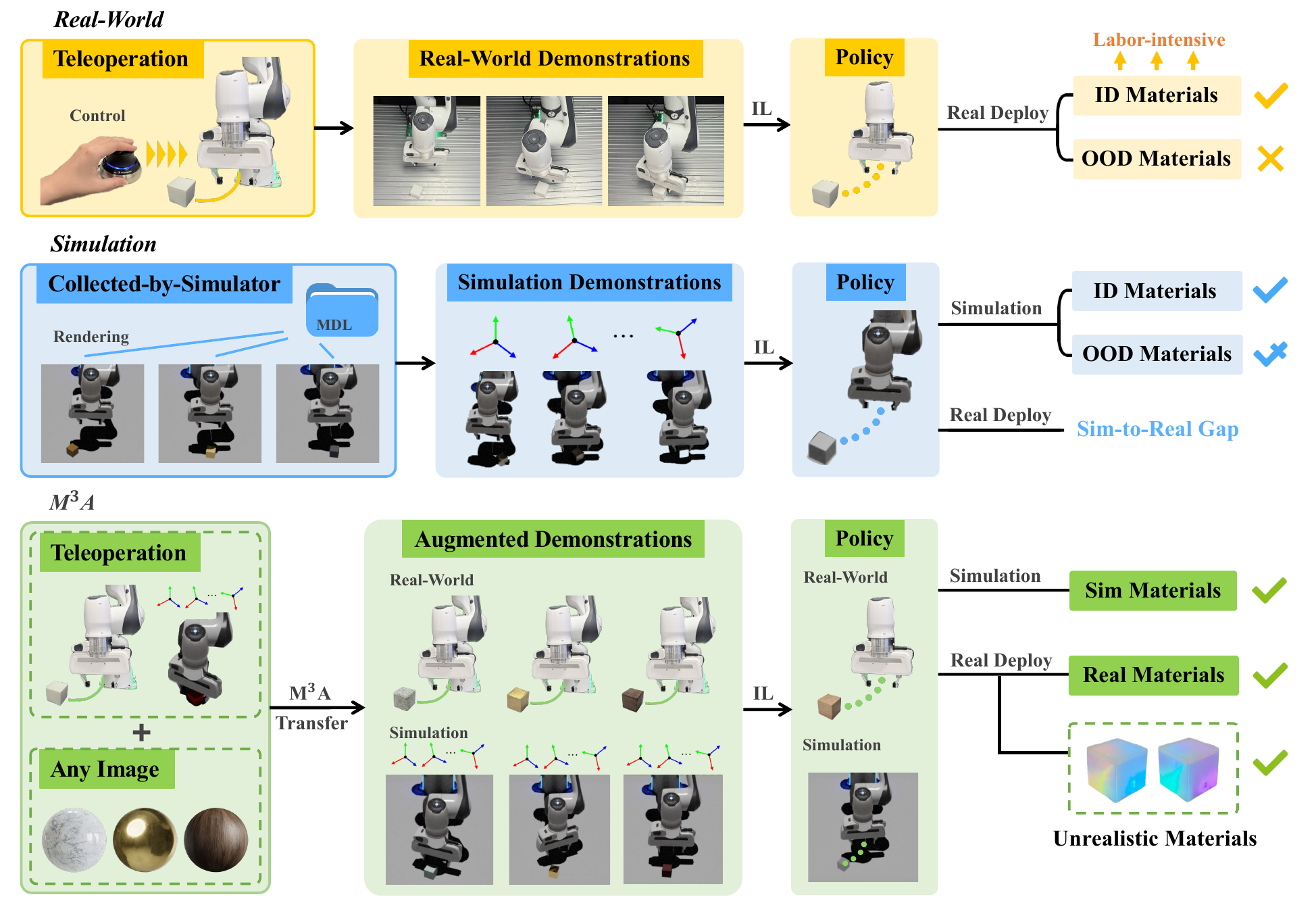}
    \caption{\textbf{Overview of the proposed $\mathbf{M^3A}$ framework}, highlighting its significant advantage in material generalization over imitation learning baselines. By synthesizing demonstrations across a wide spectrum of materials, it trains policies that robustly adapt to out-of-distribution (OOD) unseen materials and in both simulation and real-world deployment.}
    \label{fig:first_page}
\end{center}
}
]

\begin{abstract}

Material generalization is essential for real-world robotic manipulation, where robots must interact with objects exhibiting \textbf{diverse visual and physical properties}. This challenge is particularly pronounced for objects made of glass, metal, or other materials whose transparent or reflective surfaces introduce severe out-of-distribution variations. Existing approaches either rely on simulated materials in simulators and perform sim-to-real transfer, which is hindered by substantial visual domain gaps, or depend on collecting extensive real-world demonstrations, which is costly, time-consuming, and still insufficient to cover various materials.
To overcome these limitations, we resort to computational photography and introduce \textbf{Mutable Material Manipulation Augmentation (M$^3$A)}, a unified framework that leverages the physical characteristics of materials as captured by light transport for photometric re-rendering. The core idea is simple yet powerful: given a single real-world demonstration, we \textbf{photometrically re-render the scene to generate a diverse set of highly realistic demonstrations with different material properties}. This augmentation effectively decouples task-specific manipulation skills from surface appearance, enabling policies to generalize across materials without additional data collection.
To systematically evaluate this capability, we \textbf{construct the first comprehensive multi-material manipulation benchmark} spanning both simulation and real-world environments. Extensive experiments show that the M$^3$A policy significantly enhances cross-material generalization, improving the average success rate across three real-world tasks by \textbf{58.03\%}, and demonstrating \textbf{robust performance on previously unseen materials}.


\end{abstract}
\vspace{-4mm}    
\section{Introduction}
\label{sec:introduction}

Robotic manipulation has recently gained significant attention for enabling general embodied agents~\cite{zeng2018learning, qin2023dexpoint, team2023human, heng2025vitacformerlearningcrossmodalrepresentation, open6dor, black2024pi0visionlanguageactionflowmodel, open_x_embodiment_rt_x_2023, geng2023sage, zhang2024dexgraspnet20learninggenerative}, such as household robots and intelligent appliances. 
Operating in both industrial and household environments, robot agents are required to manipulate objects made of diverse materials (e.g., metal or plastic mugs), performing tasks such as grasping, placing, or pouring under varying visual and physical conditions.
Current learning-based manipulation policies~\cite{shridhar2023perceiver, zhao2023learning} mainly rely on visual perception to infer object states and guide control actions, making them highly sensitive to variations in object appearance.
In particular, the material properties of objects introduce significant appearance changes, including differences in color, surface roughness, and transparency, which lead to inconsistencies in visual perception~\cite{sajjan2020clear, li2020through, verbin2024ref}, thereby deteriorating manipulation accuracy and potentially causing physical damage.
Thus, developing embodied agents that generalize across diverse materials is essential for reliable real-world deployment.


To enhance generalization, existing methods either rely on collecting large-scale real-world demonstrations~\cite{levine2016end,kalashnikov2018scalable,pinto2016supersizing} or adopt sim-to-real transfer using simulated data and domain randomization~\cite{tobin2017domain,chebotar2019closing,andrychowicz2020learning,zhang2023efficient}.
A central challenge in material generalization is that learning robust manipulation policies would require demonstrations spanning a wide range of object materials to avoid overfitting. This requirement imposes two major limitations. First, real-world data collection becomes impractical, as acquiring diverse physical objects (e.g., wood, metal, or concrete mugs) and recording large-scale demonstrations are both labor-intensive and time-consuming~\cite{tobin2019real}. Second, while sim-to-real pipelines can easily render objects with different materials in simulation, the resulting model still suffers from visual discrepancies when transferred to the physical world due to the sim-to-real gap~\cite{zhao2020sim, wong2025survey}. This issue is further amplified for material generalization because critical visual cues, e.g., reflectance, transparency, and surface texture, are difficult to simulate with sufficient realism.

These limitations motivate us to ask: Can we develop an efficient framework for material-generalized manipulation that reduces data collection requirements while avoiding the sim-to-real gap? To this end, we propose to decouple the sources of material variation from the sources of manipulation demonstrations. Specifically, we encode material properties into compact, transferable representations that can be injected into target objects within any demonstration to alter their material appearance. This enables a single real-world demonstration to be photometrically re-rendered into numerous material variants, as long as the corresponding material representations are available. As a result, we can efficiently generate large-scale real-world mutable-material demonstrations, supporting the training and deployment of material-generalized policies without reliance on laborious data collection or imperfect simulation.

Nevertheless, the key technical challenge lies in obtaining physically plausible representations of diverse materials. Computational photography offers a principled solution to this problem~\cite{debevec2008rendering, mallick2005beyond, boivin2001image} by explicitly modeling how light interacts with surfaces. A material’s visual appearance is governed by intrinsic properties, e.g., reflectance, roughness, and translucency, that determine how incoming and outgoing light vary across illumination and viewing conditions. Traditional methods estimate these properties through photometric analysis, multi-view reflectance reconstruction, or high-dynamic-range (HDR) imaging~\cite{lin2019site,sengupta2019neural,ono2019practical,lensch2001image}, yielding spatially varying bidirectional reflectance distribution functions (BRDFs) that describe surface reflectance behavior. More recently, learning-based techniques~\cite{cheng2024zest, cheng2025marble} have enabled material editing in a physically consistent feature space, guided by depth, shading, and surface cues to generate realistic variations in color, glossiness, and transparency. These advancements provide the foundation for producing photorealistic material augmentations, thereby enabling manipulation policies to generalize robustly to previously unseen materials and bridging the visual–physical gap critical for real-world deployment.

In this paper, we propose Mutable Material Manipulation Augmentation (M$^3$A), a highly efficient framework for material-generalized manipulation policies. As shown in Fig.~\ref{fig:first_page}, we extract target objects using Grounded-SAM2~\cite{ren2024grounded} guided by the corresponding manipulation task descriptions. Given the target objects and visual appearance of certain materials, M$^3$A performs physically plausible material transformations on both real-world and simulated demonstrations. This enables a small number of collected demonstrations per task to be expanded into a large-scale, multi-material dataset without additional data collection effort. 
To systematically assess the material generalization capability of state-of-the-art policies, we construct the standard Mutable Material Manipulation (M$^3$) benchmark built on the high-fidelity Roboverse simulation platform~\cite{geng2025roboverse}. By evaluating policies in both simulation and real-world experiments, the M$^3$ benchmark ensures that methods performing well in simulation maintain consistent performance in physical environments, providing an efficient and reliable evaluation protocol.
Leveraging the diverse data generated by the M$^3$A pipeline, our learned policy exhibits strong material generalization and achieves superior zero-shot performance on unseen materials across several manipulation tasks. In summary, our contributions are threefold:
\begin{itemize}
    \item We introduce M$^3$A, a simple yet effective framework that enables physically plausible material transformations in both simulation and real-world demonstrations, supporting cross-material generalization for manipulation policies.
    
    \item We establish the M$^3$ benchmark, a comprehensive evaluation suite built on high-fidelity simulation and real-world validation, ensuring that policies performing well on the benchmark exhibit consistent capability in physical environments.

    \item Extensive experiments in both simulation and the real world show that policies trained with M$^3$A achieve strong material generalization. Our approach attains zero-shot performance on unseen materials and improves success rates by 58.03\% on average across three real-world tasks.
\end{itemize}

\section{Related works}
\label{sec:related-works}

\subsection{Data Augmentation for Robot Learning}
Data augmentation is widely used in robotic imitation learning~\cite{hussein2017imitation,radford2021learning, wei2024droma, kuang2024ramretrievalbasedaffordancetransfer} to enhance robustness without increasing real-world data collection.
Image-space augmentation methods (e.g., cropping, color jittering, random blur, and viewpoint perturbation) have been shown to improve visual robustness against lighting and camera variations, as demonstrated across several visuomotor learning methods~\cite{zhan2022learning, graf2023learning,nair2022r3m}. 
Beyond pixel-level transformations, 
geometry and physics-aware augmentation techniques exploit SE(3) pose perturbations, geometry-aware trajectory modifications, or local physics-informed transformations to increase spatial diversity while preserving action consistency~\cite{mitrano2022data,zhou2024gears,hsu2025spot}. Recently, scene-level counterfactual augmentation strategies modify distractors, backgrounds, object placements, and non-essential texture attributes to improve generalization to novel configurations and cluttered environments~\cite{ameperosa2025rocoda,dasari2019robonet}. 
These approaches collectively target variability from illumination, viewpoint, object pose, and scene composition. 

However, these methods do not explicitly address material generalization. To address this gap, recent works construct large-scale datasets with diverse material properties, including Robo360~\cite{liang2023robo360}, GAPartManip~\cite{cui2025gapartmanip}, and few-shot granular manipulation benchmarks~\cite{zhu2023few}. These datasets introduce material-level variability in simulation and real-world settings, thus providing richer training distributions for material-aware robotic manipulation. However, they remain limited in generalization to unseen tasks or novel object categories.

\subsection{Material Acquisition and Editing}
Material editing in computational photography seeks to modify surface appearance while preserving geometry, enabling visually consistent rendering under realistic illumination.
Existing methods for inverse rendering can be broadly categorized into single-image approaches that disentangle material properties from a limited observation~\cite{sengupta2019neural,chen2021invertible,idema2024neural} and those leveraging multi-view reconstruction~\cite{boss2021nerd,liu2023nero,wu2025neural}. These physically motivated pipelines have achieved high realism but were computationally demanding and sensitive to geometry and illumination accuracy, limiting their scalability for large-scale data generation.

Subsequent diffusion-based studies shifted toward semantic and generative paradigms that emphasize controllable, data-driven editing.
Single-image exemplar-based approaches~\cite{cheng2024zest,wang2024diffusion} leverage diffusion models to transfer material appearance or perform 3D editing from a single image and depth cues.
Mask-preserving methods~\cite{yin2024benchmarking,jiang2025pixelman} focus on local attribute editing while maintaining object masks or structural consistency.
Parametric and attribute-controlled frameworks~\cite{cheng2025marble, vecchio2024matfuse,zhu2024mcmat} exploit latent spaces, such as CLIP~\cite{radford2021learning} or multi-encoder representations, to manipulate fine-grained material properties including roughness, metallicity, and transparency.


\begin{figure*}[th]
    \centering
    \includegraphics[width=0.99\linewidth]{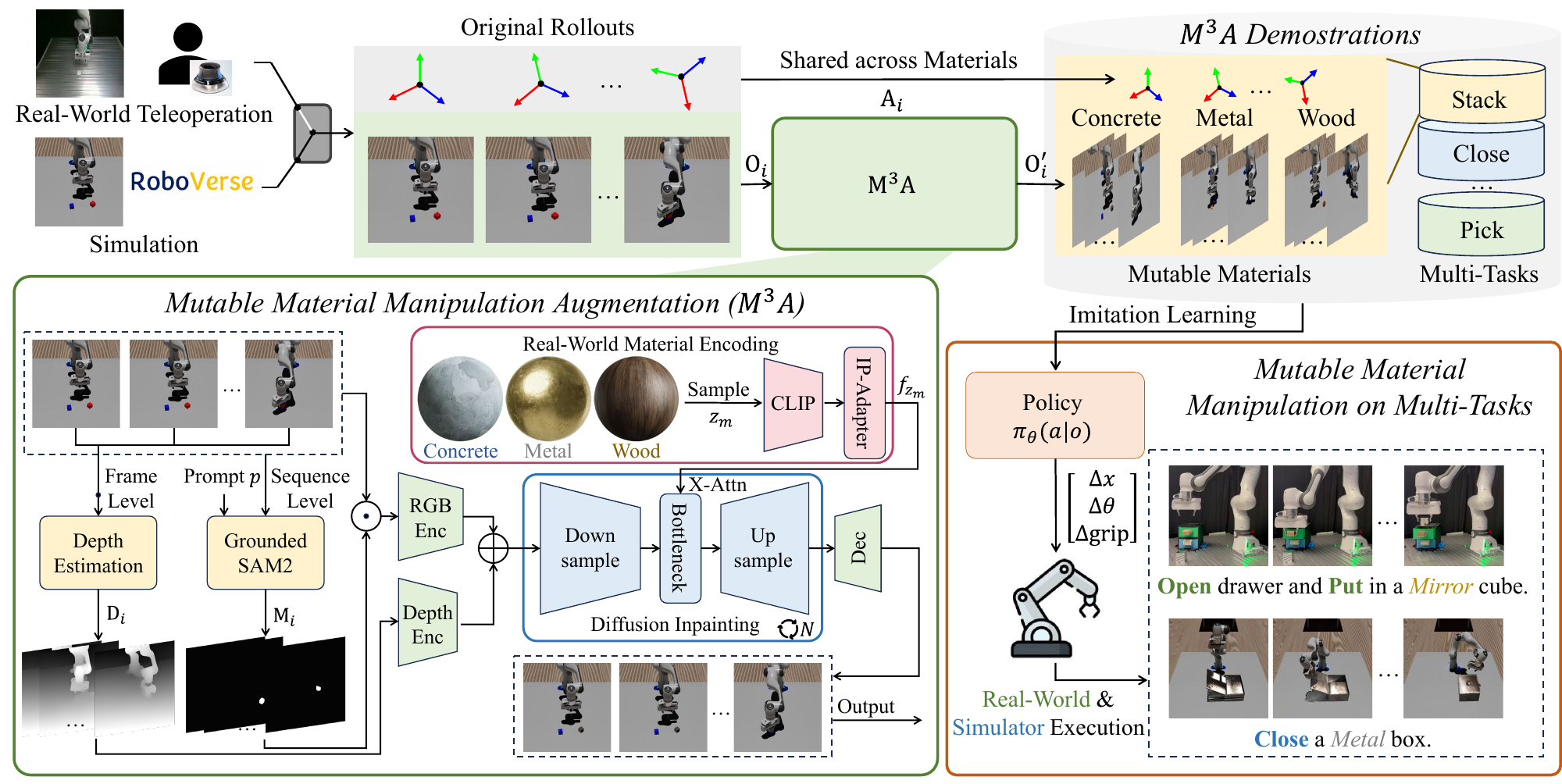}
    \caption{\textbf{The framework of M$^3$A policy.} The framework consists of three stages: (1) demonstration collection, where visuomotor trajectories (videos and action sequences) are collected from simulation or real-world environments; (2) M$^3$A, which re-composes or replaces the material appearance of manipulated objects to introduce realistic visual diversity; and (3) imitation learning, where policies are trained on the augmented demonstrations to achieve improved generalization across materials and environments.}
    \label{fig:pipeline}
\end{figure*}

\section{Method}
\label{sec:method}

\subsection{Overview}


As illustrated in~\fref{fig:pipeline}, M$^3$A provides an efficient framework for training material-generalized policies by generating physically plausible material representations and injecting them into the original demonstrations. 
Specifically, given a manipulation task, a set of demonstrations $\mathcal{D} = \{(\mathbf{O}_i, \mathbf{A}_i)\}^N_{i=1}$ are collected, containing $N$ paired demonstrations with visual observation $\mathbf{O}_i=\{o^t_{i}\}^T_{t=0}$ and the corresponding actions $\mathbf{A}_i=\{a^t_{i}\}^T_{t=0}$. 
M$^3$A augments the collected demonstrations $\mathcal{D}$, where the real-world material representations are extracted and injected into the target objects within $\mathbf{O}_i$. 
Through M$^3$A, realistic material variations, including surface reflectance, texture, and transparency, are introduced to enrich multi-material demonstrations without additional human data collection. Combining the original ($\mathcal{D}$) and augmented ($\mathcal{D'}$) demonstrations, the policy trained on M$^3$ benchmark effectively achieves the improved generalization across diverse materials.

\subsection{Mutable Material Manipulation Augmentation}
Prior studies in computational material perception~\cite{sharan2013recognizing} showed that materials can be systematically categorized based on their reflectance behavior rather than simple color or texture cues. More recently, Beveridge et al.~\cite{beveridge2025hierarchical} introduced a hierarchical representation that links local appearance patterns to global material categories, emphasizing that fine-scale reflectance and roughness jointly determine material identity. 

In robotic manipulation, material-related visual features, like reflectance, roughness, transparency, and specular highlights, hinder generalization to unseen materials when policies are trained on limited distributions.
By applying representative materials with distinct reflectance and texture profiles across broad categories, we convert each object’s single material into a diverse material set whose synthesized appearances remain photometrically close to real unseen ones. This approach reduces the discrepancy between simulated and real materials, and further enables the generation of extensive material-rich data with limited real-world data collection.

Inspired by computational photography, M$^3$A identifies specific material by its unique visual appearance under different scenarios. Subsequently, realistic augmentation can be achieved by material representations and modifications in visual feature space. Overall, the process of M$^3$A for the $i$-th demonstration can be formulated as:
\begin{equation}
    \mathbf{O}_i' = \text{M$^3$A}(\mathbf{O}_i, \mathbf{M}_i, \mathbf{D}_i, f_{z_m}),
\label{M3A_overall}
\end{equation}
where $\mathbf{O}_i$ and $\mathbf{O}'_i$ are the original and enhanced observations for the $i$-th demonstration, $\mathbf{M}_i=\{m^t_{i}\}^T_{t=0}$ and $\mathbf{D}_i=\{d^t_{i}\}^T_{t=0}$ represent the masks of the target object and depth maps for each frame, and $f_{z_m}$ denotes the representation for specific material $z_m$ from a set of material exemplars. 

Finally, the augmented demonstration set $\mathcal{D'}$ consists of enhanced observations and original actions: $\mathcal{D'} = \{(\mathbf{O}'_i, \mathbf{A}_i)\}_{i=1}^N$. Combining two demonstration sets, our M$^3$ benchmark $\mathcal{\widehat{D}}=\mathcal{D'} \cup \mathcal{D}$ enables learning material-generalized policies without additional data collection burden. In the following, we elaborate on the motivations and technical details for integrating each component.

\textbf{Mask Extraction.}
For a specific real-world manipulation task, the material typically varies only for the target object, whereas the materials of task-irrelevant objects and environments remain unchanged. To this end, we extract task-relevant foreground masks to enable the precise transfer of realistic materials to target objects.

Technically, to generate task-relevant masks, the Grounded-SAM2~\cite{ren2024grounded}, a powerful Vision-Language segmentation model, is utilized to segment target objects that are semantically grounded in the task specification. The process can be formulated as: 
\begin{equation}
    \mathbf{M}_i=\mathcal{M}(\mathbf{O}_i, p),
\end{equation} 
where $\mathcal{M}(\cdot,\cdot)$ refers to the Grounded-SAM2 to provide foreground masks $\mathbf{M}_i$, given all observations $\mathbf{O}_i$ and a task prompt $p$ for the $i$-th demonstration. The task prompt $p$ can be either a textual description (e.g., ``the red cube'') or a visual prompt (e.g., a key point or bounding box to highlight the target object). Furthermore, by taking the whole sequence $\mathbf{O}_i$ as input, the consistency of $\mathbf{M}_i$ is enhanced by referring to the correlations among observations. 

\textbf{Depth Map Estimation.} In real-world scenarios, even objects with an identical material can appear different due to the geometrical variations, such as lighting positions and shapes. To address this issue, we incorporate depth images to provide geometric priors about both the object and the environment. The geometric information enables M$^3$A to simulate material appearance variations across different scenarios, ensuring realistic multi-material augmentation. 
In simulators, physically accurate depth images are available.
However, in real-world settings, obtaining accurate depth information is challenging due to the limitations of current depth cameras towards diverse scenarios~\cite{haider2022can}. 
Alternatively, we use DPT-Hybrid (MiDaS), a depth prediction foundation model pretrained on large-scale data, to estimate robust depth images for each RGB observation:
\begin{equation}
    \mathbf{D}_i = \{\mathcal{D}(o_i^t), o_i^t \in \mathbf{O}_i\}.
\end{equation}
\textbf{Materials Transfer.} To simulate diverse material properties in the physical world, we establish an exemplar materials set $\mathbf{Z} = \{z_m\}_{m=1}^{N_z}$, where each material corresponds to a texture image $z_m$.
The CLIP vision encoder $\phi_{\text{CLIP}}(\cdot)$~\cite{radford2021learning} and an IP-Adapter $\varepsilon_{\text{IP}}(\cdot)$~\cite{ye2023ip} are then employed to extract visual features from texture images, serving as the unique representation for each material:
\begin{equation}
    f_{z_m} = \varepsilon_{\text{IP}}(\phi_{\text{CLIP}}(z_m)).
\end{equation}
As shown in Eq.~\ref{M3A_overall}, we randomly sample a material feature, $f_{z_m}$, and inject it into the bottleneck layer of a U-Net-based Stable Diffusion model~\cite{rombach2022high} to inpaint a novel material onto the target object in $\mathbf{O}_i$.
The final demonstration set is the combination of the original and augmented demonstrations with shared actions: $\mathcal{\widehat{D}}=\{(\mathbf{O}'_i, \mathbf{A}_i)\}_{i=1}^N \cup \mathcal{D}$.

Notably, the M$^3$A pipeline can convert a single target object into multiple material appearances by simply varying the reference image, $z_m$. This enables efficient scaling of material types, resulting in a comprehensive multi-material manipulation (M$^3$) benchmark. Benefiting from data diversity, policies trained on the M$^3$ benchmark are compelled to rely on material-agnostic geometric invariants (e.g., grasp points or edge contours) to perform manipulation, thereby achieving material generalization.

\subsection{Policy Training}
M$^3$A is an efficient and general augmentation pipeline that can be used as a plug-and-play module for training material-generalized policies. In this work, we focus on diffusion-based policies~\cite{chi2025diffusion}, trained under the imitation learning paradigm. Mathematically, our goal is to learn a policy $\pi_\theta(a_t|o_t)$ from the augmented demonstration set $\mathcal{\widehat{D}}$, where the $i$-th trajectory is denoted as $\tau_i = \{o'_t, a_t \}_{t=0}^T$. For simplicity, we omit the trajectory index $i$ in the following.

Diffusion-based policies formulate action prediction as a conditional denoising process over observations. During training, a random Gaussian noise $\epsilon^K$ is added to the noise-free actions $a_t$ in $\tau_i$, producing noisy actions $a_t^K$. The policy then learns to iteratively predict and remove the noise over $K$ steps to recover the original actions. Specifically, at the $k$-th iteration, the policy $\pi_\theta$ is trained to predict the added noise $\epsilon^k$ by minimizing the following objective:
\begin{equation}
\mathcal{L}_{DP} = \lVert \epsilon^k - \pi_\theta(a_t^k, k, o'_t) \rVert^2.
\end{equation}
By conditioning on the augmented observation $o'_t$, the diffusion-based policy learns action patterns that are invariant to material variations, thereby achieving material-generalized manipulation. 

\begin{figure}[t]
    \centering
    \includegraphics[width=\linewidth]{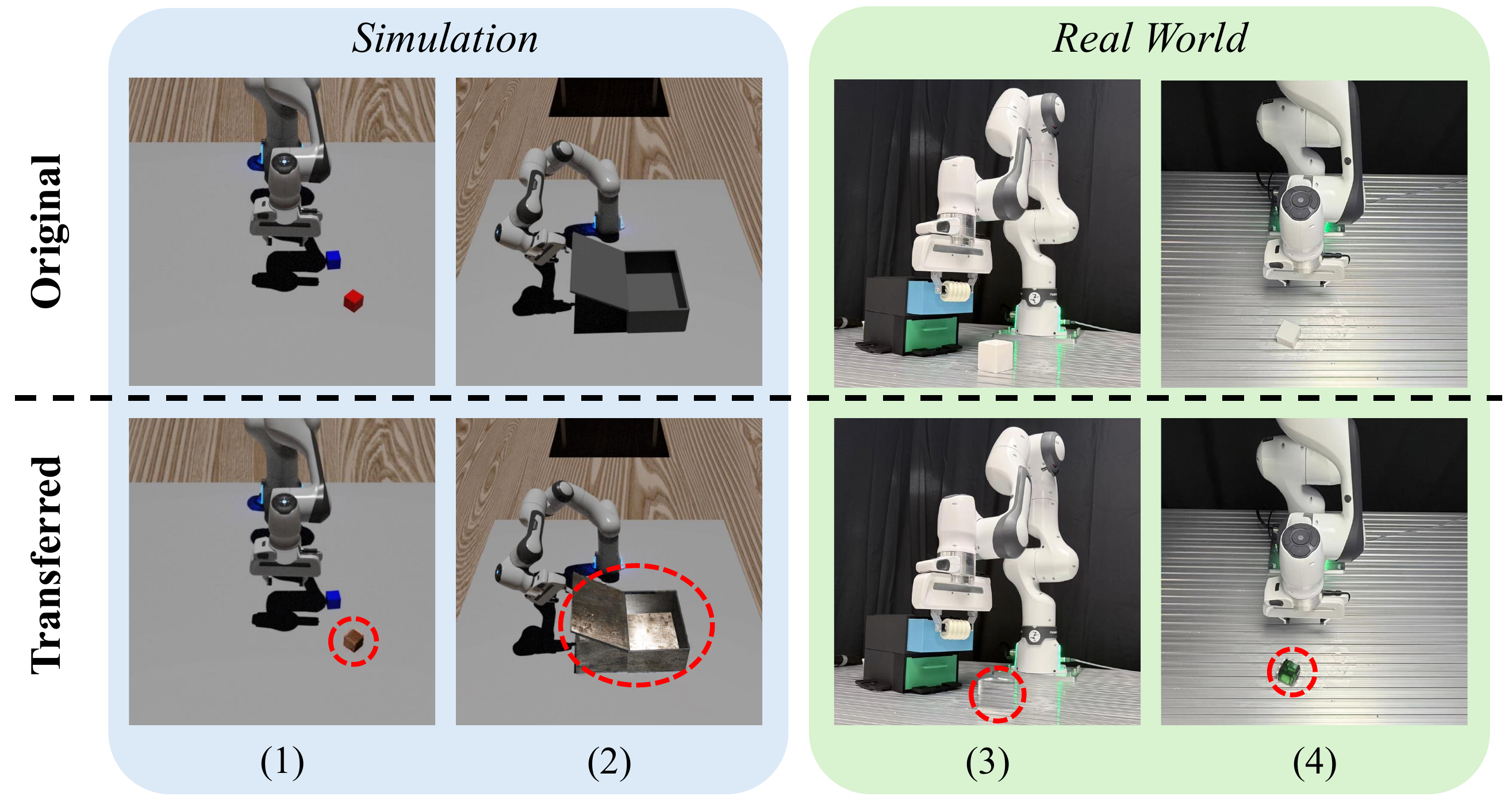}
    \caption{\textbf{Material transfer results produced by M$^3$A in both simulation and the real world.} The top row shows the original camera observations, while the bottom row presents the corresponding material-transferred outputs. The four examples illustrate: (1) red plastic to wood, (2) dark gray plastic to metal, (3) white plastic to glass, and (4) white plastic to gemstone.}
    \label{fig:material_transfer}
\end{figure}

    

\begin{table*}[t]
    \centering
    \caption{Success rates for the \textit{PickCube} task across different materials in simulation experiment.}
    \resizebox{\linewidth}{!}{
    \renewcommand{\arraystretch}{1.2}
    \setlength{\tabcolsep}{3pt}
    \scalebox{1}{
        \begin{tabular}{l|ccccccccccccc}
        \toprule
        \rowcolor{gray!15}
                Methods & Overall & Metal & Wood & Fabric & Plastic & Stone & Glass & Leather & Gems & Ceramic & Paint & Paper & Other \\
            \midrule
             DP & 11.3\% & 10.6\% & 13.8\% & 17.5\% & 11.3\% & 10.6\% & 7.5\% & 11.9\% & 10.6\% & 10.6\% & 17.5\% & 10.6\% & 10.6\% \\
             DP-Render & 21.9\% & 18.1\% & 21.9\% & 19.4\% & 21.3\% & 21.9\% & 21.3\% & 20.0\% & 20.6\% & 21.3\% & 16.9\% & 20.6\% & 18.1\% \\
             DP-M$^3$A & \textbf{34.4\% }& \textbf{30.6\%} & \textbf{36.9\%} & \textbf{31.9\%} & \textbf{29.4\%} & \textbf{33.8\%} & \textbf{27.5\% }& \textbf{27.5\%} & \textbf{24.4\%} & \textbf{33.1\% }& \textbf{31.3\%} & \textbf{32.5\% }& \textbf{31.9\%} \\
        \bottomrule
        \end{tabular}
    }
    }
    
    \label{tab:material}
\end{table*}

\subsection{Benchmark Design and Evaluation}
To fairly evaluate the material generalization capability of different policies, we establish a Mutable Material Manipulation (M$^3$) benchmark built upon RoboVerse~\cite{geng2025roboverse}, an open-source platform that supports high-fidelity robotic manipulation tasks in simulation.
As a result, the policies that achieve high performances on our benchmark can be considered to achieve comparable material generalization capability in the physical world.
Our benchmark is designed to answer two principal research questions:
\begin{itemize}
    \item Simulation Rendering vs. Computational Photography: can computational photography enhance material generalization by mitigating the sim-to-real visual gap? 
    \item Zero-shot in material domain: can a policy acquire zero-shot capability regarding materials for real-world manipulation tasks after seeing a diverse set of objects with extensive materials?
\end{itemize}
To this end, in simulation settings, the benchmark compares policies trained using conventional simulation renderings against those augmented with computational photography in M$^3$A across multiple tasks. In real-world settings, the benchmark evaluates policies trained from demonstrations involving physical objects with diverse materials and those incorporating M$^3$A, measuring their generalization ability and zero-shot performance on material domain.

\section{Experiments}
\label{sec:experiments}

To evaluate the effectiveness of the proposed M$^3$A framework in improving material generalization for robotic manipulation, we conduct comprehensive experiments in both simulation and real-world environments. The experiments are designed to assess how well one policy adapts to objects with varying material properties, such as surface texture, reflectance, and color. The primary evaluation metric is the manipulation success rate across different material domains, reflecting the policy’s generalization capability. 
For the M$^3$A implementation, we first collect demonstrations with simple baseline materials (e.g., plastic). We then apply M$^3$A transfer to the observation images of these demonstrations to generate a rich set of multi-material training data. The resulting transferred materials are illustrated in Fig.~\ref{fig:material_transfer}, demonstrating the ability of M$^3$A to produce demonstrations with realistic and diverse materials.

\subsection{Simulation experiments}
\subsubsection{Framework Overview}
All simulation experiments are conducted using the RoboVerse platform~\cite{geng2025roboverse}, which unifies a wide range of robotic manipulation tasks across multiple robotic arms and provides consistent evaluation protocols and a unified API for common simulators such as IsaacLab~\cite{mittal2023orbit} and MuJoCo~\cite{todorov2012mujoco}. We primarily employ IsaacLab for our experiments due to its high-fidelity rendering and ability to enable material randomization, both of which are essential for generating realistic material appearances and interactions.


\subsubsection{Experimental Setup}
\textbf{Task Descriptions.} In the simulation, a Franka Emika Panda robotic arm is employed to evaluate our method on three manipulation tasks to assess material generalization: 
\begin{itemize}
    \item \textit{PickCube}. This task requires the robot to pick up a textured cube. Its primary purpose is to rigorously evaluate the model's generalization to novel, unseen materials, isolating appearance variation from geometric complexity.
    \item \textit{StackCube}. This task involves picking up a cube and placing it on another. It tests the method's effectiveness in a dynamic task where visual appearance and precise placement must be coordinated.
    \item \textit{CloseBox}. This task requires closing a box lid, a motion that involves contact with a daily object, assessing the method's ability beyond simple cube manipulation.
\end{itemize}

\textbf{Benchmarking Policies.} We collect expert demonstration trajectories from three distinct sources to train three policies:
(1) DP. Demonstrations containing objects with default materials given by RoboVerse. 
(2) DP-Render. Demonstrations from the original environment, modified with varied material and lighting conditions through RoboVerse to increase basic visual diversity. 
(3) DP-M$^3$A. Demonstrations produced by our M$^3$A framework, which transfers realistic materials to the manipulated objects while strictly preserving the motion trajectory consistency.

\textbf{Training Configuration.}
All three policies are trained using DP~\cite{chi2025diffusion}. We train all policies for 150 epochs using a learning rate of $1\times 10^{-4}$ and Adam optimizer~\cite{kingma2014adam}. 

\subsubsection{Experimental Results}
\textbf{Material-wise Generalization.}
For the PickCube task, all materials are first grouped into twelve categories.
A total of 160 trajectories collected within the RoboVerse environment are used as base demonstrations to provide action labels for training three benchmark policies.
We then evaluate each policy’s performance separately on each material category.
Besides, the overall performance is computed on a fixed set of materials, including samples from all categories.


From the quantitative results in~\tref{tab:material}, the proposed M$^3$A policy outperforms the other methods, DP and DP-Render.
The original DP exhibits notable performance degradation on the material categories with specular reflections or complex textures, revealing a critical dependency on the appearance characteristics.
Notably, while the Rendered baseline provides a marginal average improvement by introducing basic visual variability, its gains are inconsistent and fail to generalize robustly across all material types. 
In contrast, the proposed DP-M$^3$A framework achieves superior success rates in every material category, increasing about 12.5\% success rate than DP-Render. 
This consistent performance uplift, especially on challenging materials like metals and transparent surfaces, demonstrates that the computational photography technology can improve the accuracy of robotic policy due to the more realistic material appearances than those rendered from simulators. 
The results confirm that the proposed M$^3$A is effective in improving material generalization capability of robotic policy.

\textbf{Evaluation on Manipulation Tasks.}
The effectiveness of our M$^3$A framework extends beyond material-specific generalization to enhance robustness across diverse manipulation tasks, as summarized in \tref{tab:across task}. On both the StackCube and CloseBox tasks, which involve dynamic multi-object interaction and articulation, policies trained with our augmented demonstrations consistently outperform those trained on original data. The performance improvement is particularly significant as these tasks integrate geometric, spatial, and physical reasoning alongside visual perception.
By exposing the policy to the diverse realistic material appearances during training, the policy focuses more on task-relevant geometric and physical features, rather than overfitting to specific visual correlations.

\begin{table}[t]
    \centering
    \renewcommand{\arraystretch}{1.2}
    \setlength{\tabcolsep}{3.5pt}
    \caption{Comparison of success rates between DP and our M$^3$A method across three simulated manipulation tasks.}
    \scalebox{0.9}{
        \begin{tabular}{l|cccc}
        \toprule
            Methods & Average& PickCube & CloseBox & StackCube\\
            \midrule
             DP & 10.16\%& 11.3\% & 16.7\% & 2.5\%\\
             DP-M$^3$A & \textbf{22.80\%}& \textbf{34.4\%} & \textbf{27.1\%} & \textbf{6.9\%} \\
        \bottomrule
        \end{tabular}
    }
    
    \label{tab:across task}
\end{table}

\subsection{Real World Experiments}
\begin{table*}[t]
    \centering
    \renewcommand{\arraystretch}{1.1}
    \setlength{\tabcolsep}{3pt}
    \caption{Comparison of real-world performance across three cube-manipulation tasks involving eleven material types.}
    \resizebox{\linewidth}{!}{ 
    \begin{tabular}{l|cccccccccccc}
    \toprule
        
        Methods & Average&\underline{White}& \underline{Beech} & \underline{Rubber}& \underline{Wool}& \underline{Silk}& \underline{Foam}& Glass& Mirror& Walnut &Leather& Flash\\
        \midrule
\rowcolor{gray!15}
\multicolumn{13}{l}{\textit{Picking Task}} \\
        \midrule
         DP& 22.35\%& \textbf{100.0\%}& 4.2\%& 12.5\%& 45.8\%& 12.5\%& 0.0\%& 4.2\%& 37.5\%& 8.3\%& 4.2\%& 16.7\%\\
         DP-6& 48.86\% & 87.5\%& \textbf{87.5\%}& 62.5\%& 62.5\%& 75.0\%& \textbf{87.5\%}& 12.5\%& 25.0\%& 12.5\%& 12.5\%& 12.5\%\\
         DP-M$^3$A & \textbf{89.40\%}& 95.8\%& 66.7\%& \textbf{100.0\%}& \textbf{100.0\%}& \textbf{100.0\%}& \textbf{87.5\%}& \textbf{75.0\%}& \textbf{100.0\%}& \textbf{79.2\%}& \textbf{79.2\%}& \textbf{100.0\%} \\
        \midrule
        \rowcolor{gray!15}
        \multicolumn{13}{l}{\textit{Picking \& Placing Task}} \\
        \midrule
         
         DP& 30.68\%& \textbf{100.0\%}& 0.0\%& \textbf{100.0\%}& \textbf{100.0\%}& 0.0\%& 0.0\%& 0.0\%& 0.0\%& 0.0\%& 0.0\%& 37.5\%\\
         DP-6 &59.09\% & \textbf{100.0\%}& \textbf{87.5\%}& \textbf{100.0\%}& 87.5\%& 37.5\%& 62.5\%& \textbf{75.0\%}& 0.0\%& 25.0\%& 0.0\%& 75.0\%\\
         DP-M$^3$A &\textbf{68.18\%}& 100.0\%& \textbf{87.5\%}& 87.5\%& 87.5\%& \textbf{75.0\%}& \textbf{87.5\%}& 37.5\%& \textbf{12.5\%}& \textbf{25.0\%}& \textbf{50.0\%}& \textbf{100.0\%}\\
        \midrule
        \rowcolor{gray!15}
        \multicolumn{13}{l}{\textit{Long-horizon Picking \& Placing Task}} \\
        \midrule
         DP&{24.24\%}& \textbf{91.7\%}& 8.3\%& 58.3\%& 25.0\%& 41.7\%& 0.0\%& 0.0\%& 8.3\%& 0.0\%& 0.0\%& 33.3\%\\
         DP-6&{57.57\%}& \textbf{91.7\%}& 66.7\%& \textbf{91.7\%}& 66.7\%& 83.3\%& \textbf{100.0\%}& 0.0\%& 83.3\%& 0.0\%& 8.3\%& 41.7\%\\
         DP-M$^3$A &\textbf{93.94\%}& \textbf{91.7\%}& \textbf{100.0\%}& \textbf{91.7\%}& \textbf{83.3\%}& \textbf{91.7\%}&83.3\%& \textbf{100.0\%}& \textbf{91.7\%}& \textbf{100.0\%}& \textbf{100.0\%}& \textbf{100.0\%}\\
    \bottomrule
    \end{tabular}
    }
    \label{tab:real_world_experiment}
\end{table*}
\subsubsection{Experimental Setup}

\begin{figure}[th]
    \centering
    \includegraphics[width=\linewidth]{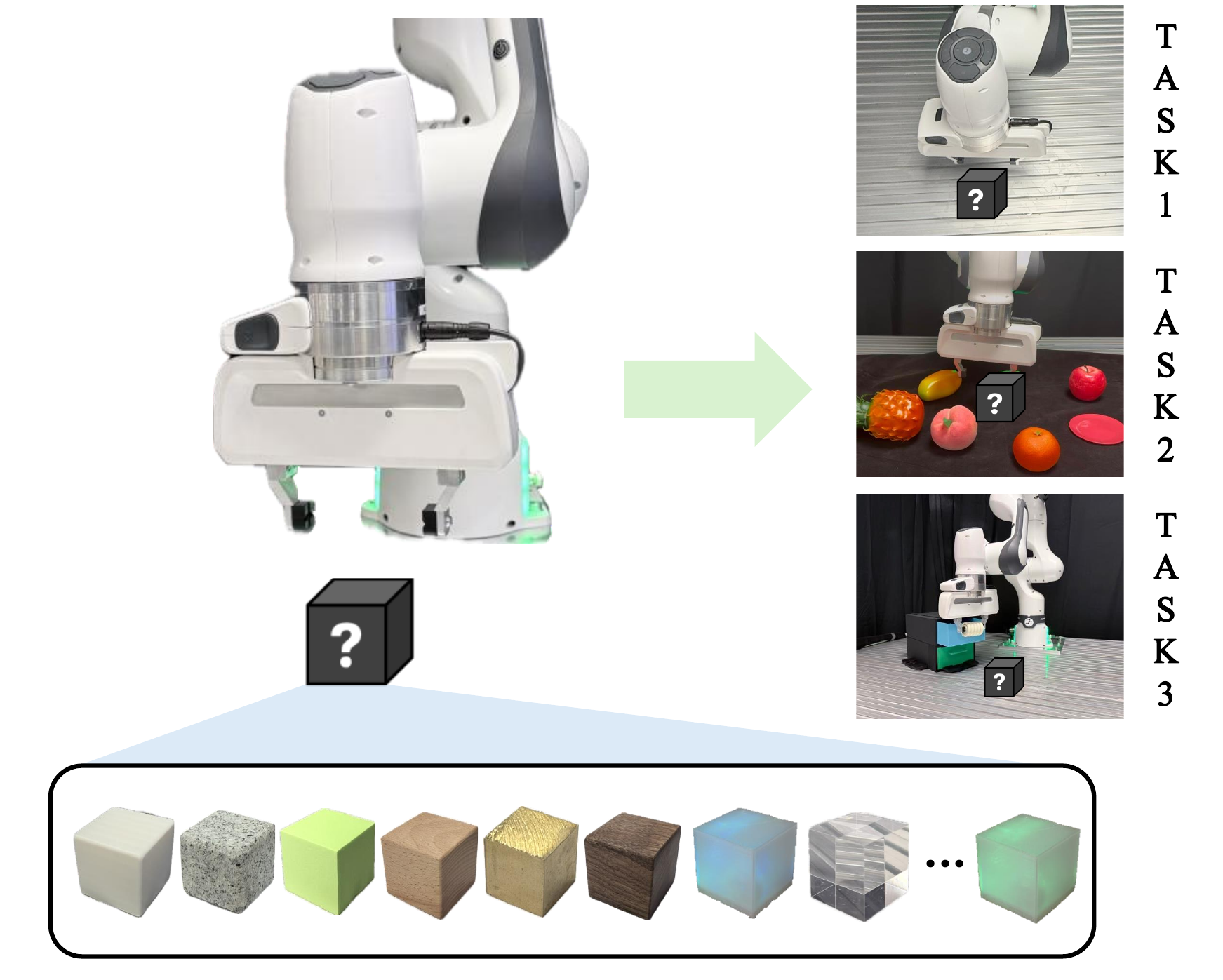}
    \caption{\textbf{Real-world experiment settings.} The FR3 manipulates cubes with eleven different materials to finish three tasks: (1) Picking, (2) Picking \& placing, (3) Long-horizon picking \& placing. }
    \label{fig:real_world_settings}
\end{figure}

For real-world experiments, as shown in~\fref{fig:real_world_settings}, we use $5\times5\times5$ cm cubes with 11 diverse materials in three robotic manipulation tasks, enabling the evaluation of material generalization under consistent geometry. 

\textbf{Task Descriptions.}
The details of three real-world tasks: (1) \textit{Picking}, (2) \textit{Picking \& Placing}, and (3) \textit{Long-Horizon Picking \& Placing}, are elaborated as follows:
\begin{itemize}
    \item \textit{Picking}. The robotic arm picks a cube of specific material from random positions with a clean background.
    \item \textit{Picking \& Placing}. The robotic arm picks a cube of specific material and places them into the target plate with a messy environment with distractors.
    \item \textit{Long-horizon Picking \& Placing}. The robotic arm first grasps and opens a drawer, picks a cube of a specific material from the table, and places it inside the drawer.
\end{itemize}

\textbf{Experimental Configurations.}
For the hardware, a Franka Emika Research 3 (FR3) and two RealSense D455 cameras are employed for both demonstration collection and manipulation. 
For the software, we follow the configuration in HIL-SERL~\cite{luo2025precise}, and run the control system on a PC equipped with an NVIDIA RTX 5080 GPU (16 GB).

\textbf{Benchmarking Policies.} Three kinds of DP are compared: (1) DP is trained only with a white plastic cube in demonstrations, (2) DP-6 is trained using demonstrations with cubes of six materials, and (3) our DP-M$^3$A is trained with demonstrations augmented by the proposed M$^3$A.
Notably, all material images used for augmentation in M$^3$A are collected from the web and exhibit discrepancies compared to their real-world visual appearances.
Thus, the performance of DP (M$^3$A) in real-world settings reflects its zero-shot capability on materials in the physical environment.

\subsubsection{Experimental Results}
The real-world experimental results are summarized in \tref{tab:real_world_experiment}. 
Specifically, the underlined \underline{materials} indicate those used in demonstrations to train DP-6 while the remaining materials are unseen during DP-6 training.
The proposed M$^3$A strategy substantially enhances generalization on materials, achieving the highest average success rates of 89.40\%, 68.18\%, and 93.94\% in the respective tasks.

\textbf{Picking.} Both DP and DP-6 perform well on seen materials, attaining 100\% and 75\% success rates, respectively. However, their performance drops sharply on unseen materials, reaching only on an average of 15\% on materials excluding white for DP and and 15\% on unseen materials for DP-6. 
The significant drop of success rate presents their weakness in material generalization capabilities.
In contrast, the proposed DP-M$^3$A maintains consistent performance across all materials, achieving more than 75.0\% success rate in most of the materials, despite being trained solely on augmented data collected from online materials instead of the data collected in the real world.
This demonstrates the strong generalization ability of the M$^3$A framework.

\textbf{Picking \& Placing.} The robotic manipulation faces the problems of distractors, limiting the performance of DP. As we can see from the table, DP and DP-6 fail to pick cubes in some materials at all, with 0\% success rate. 
However, after augmenting the demonstrations, the DP-M$^3$A can achieve manipulating the cubes with these materials, reflecting the effectiveness of the proposed M$^3$A framework.
However, the performance of Task 2 is not as high as that of Task 1. This may result from the imprecise mask prediction and depth estimation due to the clustered environments.

\textbf{Long-horizon Picking \& Placing.} The policies face the substantial challenges for DP and material augmentation due to their accumulated errors and strong temporal dependencies. 
Remarkably, however, the DP-M$^3$A achieves an overall 93.94\% success rate on all provided materials, outperforming the origin DP and DP-6, validating the effectiveness and robustness of our method.
For the DP and DP-6, the success rate drops (91.7\% to 17.5\% and 83.4\% to 26.7\%) also happen in this tasks.   

In the real-world experiments, the traditional DP algorithm performs reliably on objects with seen materials but shows clear limitations when encountering unseen materials.
By contrast, the proposed M$^3$A framework significantly enhances material generalization through realistic computational-photography rendering.
Notably, even though M$^3$A relies solely on online material images, the resulting DP-M$^3$A policy still achieves strong performance on real-world objects, exhibiting a clear zero-shot capability. These findings demonstrate that computational-photography-based material augmentation can effectively transfer to the real world and equip robotic policies with robust zero-shot material generalization.

\section{Conclusion}
\label{sec:conclusion}
In this work, we present a unified framework for material-generalized robotic manipulation, bridging the gap between visual diversity and task adaptability. By drawing inspiration from computational photography, we introduce a material editing mechanism that effectively decouples manipulation skills from material appearances, enabling efficient augmentation of imitation learning data. Furthermore, we establish a systematic benchmark to evaluate cross-material generalization and verify our approach across both simulated and real environments. Extensive results demonstrate that our method achieves substantial gains in success rate and robustness, particularly on unseen materials, highlighting its potential for scalable and material-agnostic robotic learning in the real world.

However, the proposed method still has limitations. In real-world settings, we observe that the accuracy and consistency of material transfer are influenced by the mask quality, particularly in messy environments with a cluster of distractors. In the future, we aim to improve mask prediction to solve this issue.

{
    \small
    \bibliographystyle{ieeenat_fullname}
    \bibliography{main}
}

\clearpage
\setcounter{page}{1}
\maketitlesupplementary

\section*{Supplementary Experiment}
\label{sec:supp exp}

\begin{figure}[th]
    \centering
    \includegraphics[width=\linewidth]{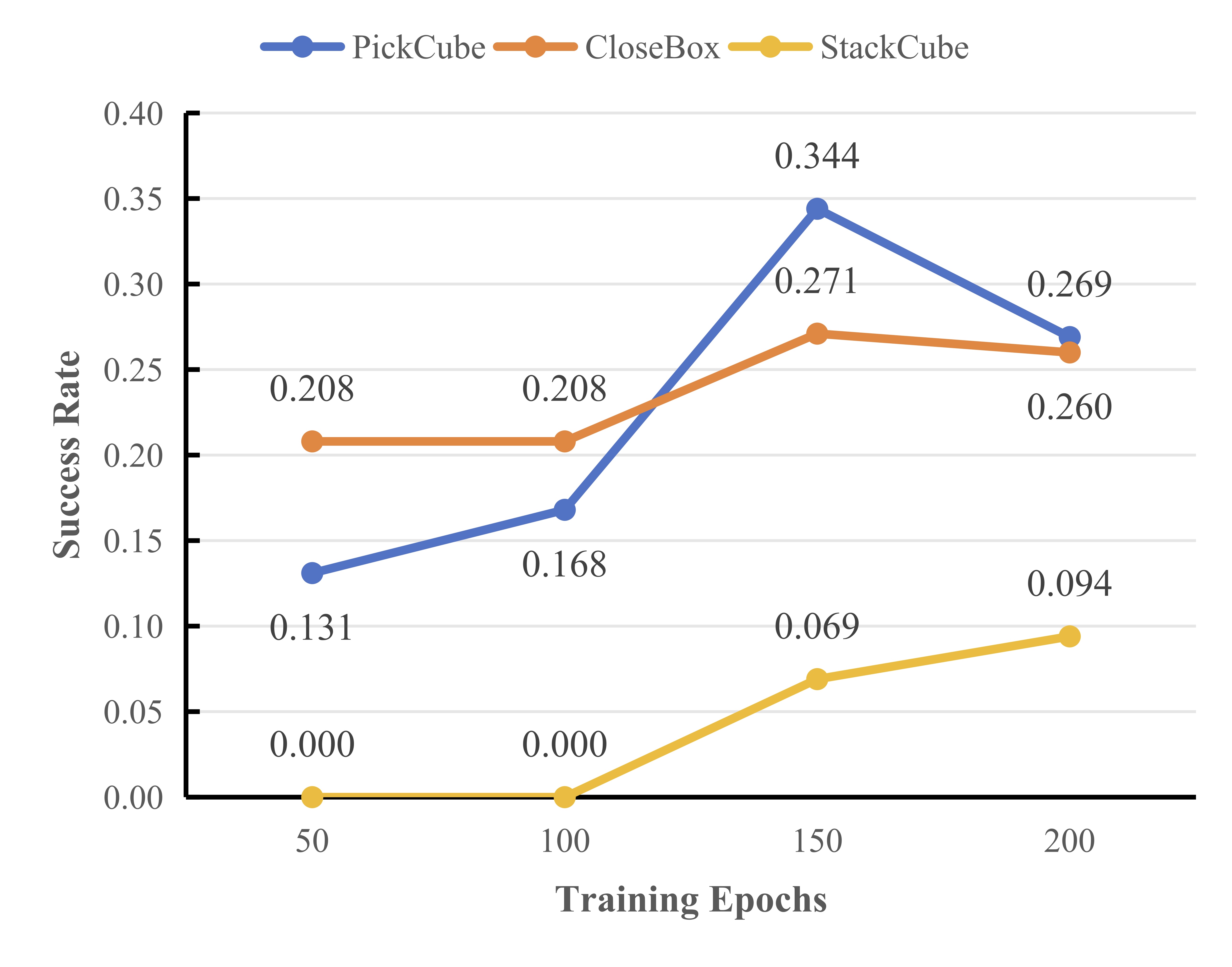}
    \caption{\textbf{Success rate of simulation tasks under varying DP training epochs.} }
    \label{fig:sr_epochs}
\end{figure}

As a supplementary analysis to the simulation experiments, we evaluated the performance of the DP-M$^3$A method across different training epochs, as shown in the figure\ref{fig:sr_epochs}. 

In the PickCube and CloseBox tasks, the performance improved with increasing training epochs up to 150 epochs. However, after 150 epochs, additional training resulted in a decrease in success rate. For the StackCube task, the success rate was 0 for fewer than 100 training epochs, but as the number of epochs increased, the success rate improved, reaching higher levels within 200 epochs. This difference across tasks may be due to the higher complexity of the StackCube task compared to the PickCube and CloseBox tasks, where fewer training epochs are insufficient for the robotic arm to learn the necessary features and strategies effectively.

\section*{Experiment Videos}
For all the simulations and real-world experiments mentioned in the paper, we provide corresponding video files that demonstrate the successful execution of the tasks, showcasing the effectiveness of the M$^3$A method across different scenarios.

\textbf{Simulation Tasks.}
For the simulation tasks, we offer the following video files, each demonstrating the successful execution of the tasks under three different kinds of materials:
\begin{itemize}
    \item \texttt{CloseBox\_simulation.mp4}: A silent video showing the CloseBox task.
    \item \texttt{PickCube\_simulation.mp4}: A silent video displaying the PickCube task.
    \item \texttt{StackCube\_simulation.mp4}: A silent video illustrating the StackCube task.
\end{itemize}

\textbf{Real-World Experiments.}
Similarly, for the real-world experiments, we provide video files that show the successful execution of tasks on physical cubes made of multiple materials:
\begin{itemize}
    \item \texttt{picking.mp4}: A silent video demonstrating the execution of the Picking task.
    \item \texttt{picking\_and\_placing.mp4}: A silent video showcasing the performance in the Picking \& Placing task.
    \item \texttt{long-horizon.mp4}: A silent video illustrating the process of the Long-horizon Picking \& Placing task.
\end{itemize}

\end{document}